%% file: main.tex
\documentclass[sigconf]{acmart}

\usepackage{tikz}
\usepackage{amsmath,bm}
\usepackage{subfiles}
\usepackage{multicol}
\usepackage{framed}
\usepackage{kantlipsum}
\usepackage{lipsum}
\usepackage{setspace}
\usepackage{amsfonts}
\usepackage{enumitem,kantlipsum}
\usepackage[english]{babel}
\usepackage{mdframed}
\usepackage{lipsum}
\usepackage{xcolor}
\usepackage{makecell}
\usepackage{multirow}
\usepackage{float}
\usepackage{algorithm}
\usepackage{algpseudocode}

\newmdtheoremenv{definition}{Definition}
\usepackage{booktabs,caption}
\usepackage[flushleft]{threeparttable}
\usepackage{bm}

\copyrightyear{2026}
\acmYear{2026}
\setcopyright{cc}
\setcctype{by}
\acmConference[DAC '26]{63rd ACM/IEEE Design Automation Conference}{July 26--29, 2026}{Long Beach, CA, USA}
\acmBooktitle{63rd ACM/IEEE Design Automation Conference (DAC '26), July 26--29, 2026, Long Beach, CA, USA}
\acmDOI{10.1145/3770743.3804227}
\acmISBN{979-8-4007-2254-7/2026/07}

\makeatletter
\let\ACM@orig@mkbibcitation\@mkbibcitation
\def\@mkbibcitation{\let\textsuperscript\@gobble\ACM@orig@mkbibcitation}
\renewcommand{\authornote}[1]{%
  \g@addto@macro\@authornotes{%
    \begingroup\let\@makefnmark\@empty\footnotetext{#1}\endgroup}}
\makeatother

\begin{document}

\title{Accelerating GPU Inference of Large Language Models with Moderately Unstructured Sparse Weight Matrices}



\author{Tao Lu\textsuperscript{\S,*}, Haoyu Wang\textsuperscript{\P,*}, Zonghui Wang\textsuperscript{\P,\dag}, Keshen Xiang\textsuperscript{\P}, Jiaheng Zhang\textsuperscript{\S,\dag}, Wenzhi Chen\textsuperscript{\P}}
\authornote{\textsuperscript{*}Equal contributions. \textsuperscript{\dag}Corresponding authors.}
\affiliation{%
  \institution{\textsuperscript{\S}National University of Singapore, \textsuperscript{\P}Zhejiang University}
  \country{Singapore, China}}
\email{{lutao, jhzhang}@nus.edu.sg, {whaoyu, zhwang, keshen, chenwz}@zju.edu.cn}

\pagenumbering{gobble}

\renewcommand{\shortauthors}{Tao Lu et al.}

\subfile{text/abstract}

\maketitle
\pagestyle{plain}

\section{Introduction}

\subfile{text/intro}

\vspace{-0.2cm}
\section{Background}
\vspace{-0.1cm}

\subfile{text/background}


\vspace{-0.2cm}
\section{Storage Format Design}
\subfile{text/storage}

\section{SpMM Kernel Design}
\subfile{text/kernel}

\vspace{-0.2cm}
\section{Evaluation}
\vspace{-0.1cm}
\subfile{text/evaluation}

\section{Limitation and Discussion}

\subfile{text/limitation}

\section{Conclusion}
\subfile{text/conclusion}

\bibliographystyle{ACM-Reference-Format}
\bibliography{ref}

\end{document}

%% file: text/abstract.tex
\begin{abstract}

With the growing deployment of large language models (LLMs), LLM inference cost has become a key challenge. Pruning techniques that introduce sparsity into weight matrices can accelerate inference. However, maintaining model quality typically limits pruning to moderate unstructured sparsity (around 50\%). At these sparsity levels, none of the existing GPU kernels for sparse matrix multiplication (SpMM) can outperform their dense counterparts. This paper proposes an efficient GPU inference method for LLMs with moderate sparsity. We propose a three-layer matrix storage format comprising: (i) a Sparse-TC layer enabling sparse tensor cores to accelerate SpMM; (ii) a Slot-Filling layer using parallel differential distance for matrix compression while supporting low-cost on-chip decoding; (iii) a lightweight Residual Layer ensuring correct SpMM computation. Building on this format, we design a SpMM kernel that jointly utilizes sparse tensor cores and CUDA cores. This design enables an efficient execution pipeline and overlaps on-chip computation with memory access. Evaluations show that our work is the first to outperform dense matrix multiplication on modern GPUs equipped with high-bandwidth memory (HBM). It achieves up to 1.64× kernel-level speedup over SpInfer (EuroSys'25, Best paper) and up to 1.41× end-to-end speedups over FlashLLM (VLDB'24). Our source code: https://github.com/moui0/cudac.

\end{abstract}

%% file: text/intro.tex
With the growing adoption of large language models (LLMs) \cite{gpt2, attention, opt, deepseek} in applications such as natural language understanding, text generation, and code completion, the high computational cost of LLM inference has become a critical concern. To mitigate this, recent works, such as SparseGPT \cite{sparseGPT}, Wanda \cite{wanda}, and RIA \cite{ria}, have proposed effective pruning strategies that remove less important values from the weight matrices. These approaches aim to accelerate inference while preserving model quality.

However, unlike small-scale models, where high sparsity can be introduced with minimal impact on the model quality \cite{prune_trans, sparRNN}, LLMs are significantly more sensitive to high sparsity levels, which can severely degrade inference quality. As shown in Table \ref{tbl:perplexity}, state-of-the-art pruning schemes \cite{ria, wanda} achieve good performance in LLMs only when the weight matrices have \textbf{moderate unstructured sparsity}, typically around 50\%. When the sparsity level exceeds 70\%, the model's perplexity score, a key metric for evaluating the quality of language model outputs, increases sharply, which indicates a substantial deterioration in the model's ability to generate coherent and contextually appropriate responses.

Although moderate unstructured sparsity is highly effective in preserving model quality, a major bottleneck lies in the fact that none of the existing GPU kernels \cite{sparta, sputnik, highsc, tiled_csr, dtc, Magicube, qgtc, sparRNN, sparsetir, tcgnn, spinfer} for sparse matrix multiplication (SpMM) can outperform their dense counterparts, such as cuBLAS \cite{cublas}, the standard library for dense matrix multiplication. This performance gap indicates that, despite pruning reducing the number of non-zero weights, the speedup achieved with current sparse GPU kernels is negative.

\begin{table}[t] \caption{Perplexity scores of prominent LLM pruning schemes, Wanda \cite{wanda} and RIA \cite{ria}, evaluated across various sparsity levels and topologies on the WikiText-2 dataset, where a lower perplexity score indicates better performance.}
\label{tbl:perplexity}

\begin{threeparttable}
\centering
\renewcommand\arraystretch{0.7}{
\setlength{\tabcolsep}{1mm}{
\begin{tabular}{p{1.1cm}<{\centering} | p{1.7cm}<{\centering} | p{1.1cm}<{\centering} | p{1.2cm}<{\centering} p{1.2cm}<{\centering} p{0.9cm}<{\centering}}

\Xhline{1pt}

\multirow{2}*{Sparsity} & \multirow{2}*{Topology}& \multirow{2}*{Scheme} & Llama & Llama& Opt \\

~ & ~ & ~ & 7B & 13B & 1.3B \\

\cline{1-6}
\normalsize Dense & - & - & 5.47 & 4.88 & 14.6   \\
\cline{1-6}

\normalsize \multirow{4}*{50\%} & \small \multirow{2}*{unstructured} & Wanda & 7.79 & 6.28 & \textbf{18.5} \\
\cline{3-6}
\normalsize ~ & ~ & RIA & \textbf{6.88} & \textbf{5.95} &  18.9 \\
\cline{2-6}
\normalsize ~ & \multirow{2}*{2:4} & Wanda & 11.6 & 9.69 & 28.3 \\
\cline{3-6}
\normalsize ~ & ~ & RIA & 11.3 & 8.44 & 27.4\\
\cline{1-6}
\normalsize \multirow{2}*{60\%} & \small \multirow{2}*{unstructured} & Wanda & 15.3 & 9.63 & 38.8 \\
\cline{3-6}
\normalsize ~ & ~ & RIA & 10.4 & 7.84 & 26.2 \\
\cline{1-6}

\normalsize \multirow{2}*{70\%} & \small \multirow{2}*{unstructured} & Wanda & 214.9 & 105.0 & 231.2 \\
\cline{3-6}
\normalsize ~ & ~ & RIA & 68.8 & 52.0 & 98.5 \\
\Xhline{1pt}

\end{tabular}

}}

\end{threeparttable}
\end{table}

Unlike SpMM kernels for highly sparse matrices, accelerating SpMM under moderate unstructured sparsity is far more challenging. \textbf{Challenge 1: Tensor core incompatibility}. High-efficiency tensor cores on modern GPUs only support structured 2:4 sparsity, making them incompatible with unstructured pruning where nonzeros are irregularly distributed. Thus, they cannot directly accelerate SpMM in such settings. \textbf{Challenge 2: High proportion of metadata overhead}, traditional storage formats, such as Compressed Sparse Row (CSR) \cite{csr}, require storing full positional metadata for each non-zero element. At moderate sparsity levels, particularly around 50\%, the positional metadata can occupy as much space as the non-zero elements themselves. \textbf{Challenge 3: High on-chip decoding overheads in matrix compression.} While matrix compression techniques such as bitmap-based encoding \cite{spinfer} effectively reduce metadata storage overhead, they incur additional on-chip decoding workloads. Because the large volume of metadata must be decoded on CUDA cores instead of on faster tensor cores, the decoding process cannot keep up on modern GPUs with high-bandwidth memory (HBM), where on-chip computation cannot be fully overlapped by global memory access.

\begin{figure}[t]
\centering
\includegraphics[scale=0.26]{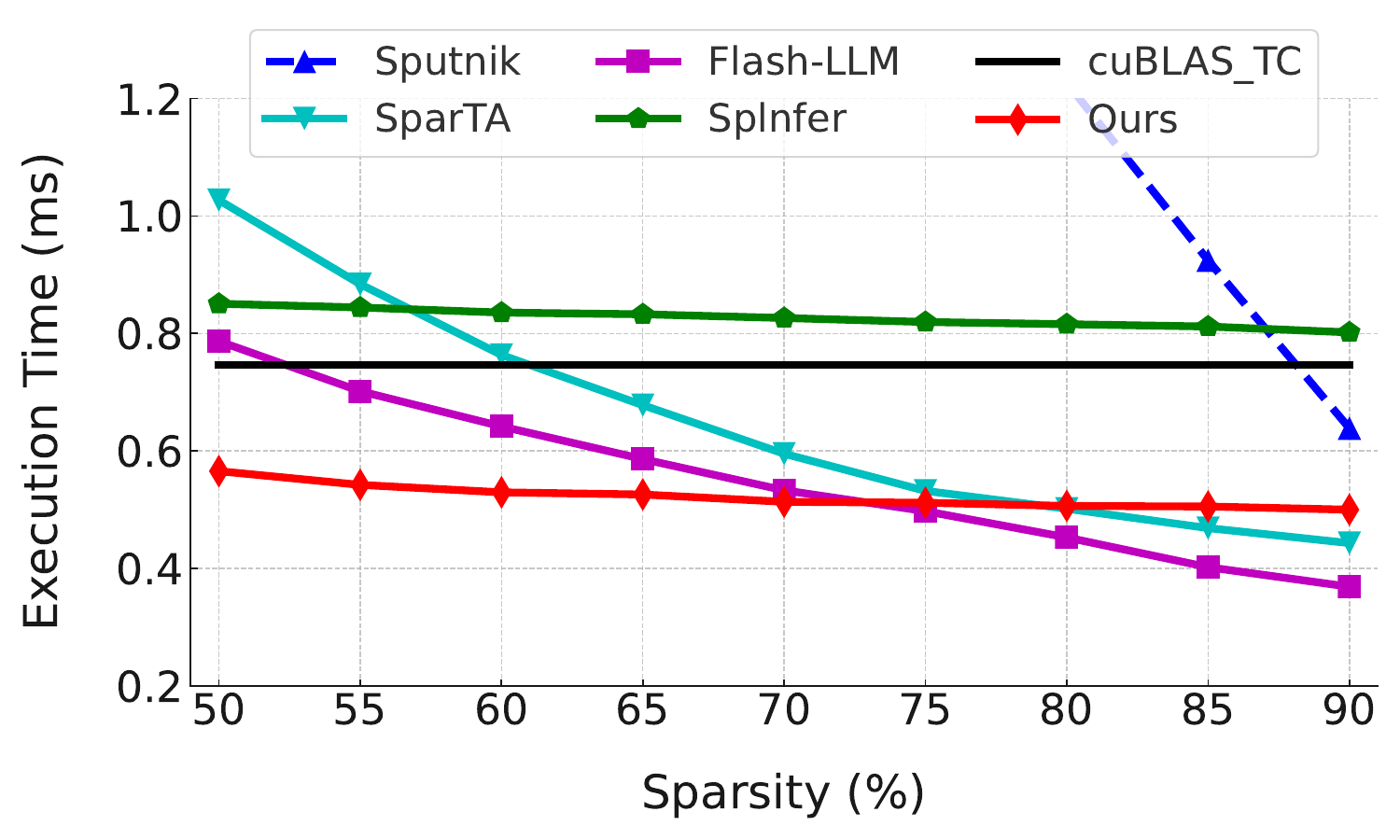}
\caption{Execution time of unstructured SpMM implementations against cuBLAS on NVIDIA H100 GPU equipped with HBM. (M/K/N = 66K/16K/32, typical in LLM inference)}
\label{fig:sparisty}
\end{figure}

In this paper, we accelerate GPU inference for LLMs with moderately unstructured sparse weight matrices. Our approach is designed to address the three fundamental challenges outlined above. On one hand, we propose a three-layer matrix storage format comprising: (i) a Sparse-TC layer enabling sparse tensor cores to accelerate SpMM; (ii) a Slot-Filling layer using parallel differential distance for matrix compression while supporting low-cost on-chip decoding; (iii) a lightweight Residual Layer ensuring correct SpMM computation. Building on this format, we design a co-optimized SpMM kernel that jointly utilizes sparse tensor cores and CUDA cores, leveraging the former to accelerate structured matrix operations and the latter to flexibly manage the irregular decoding for our storage format. This design enables an efficient execution pipeline and overlaps on-chip computation with memory access. 

The following is a summary of our contributions:

\begin{itemize}

\item We propose a sparse matrix storage format tailored for moderate unstructured sparsity, addressing three challenges in sparse LLM inference on GPUs: enabling compatibility with sparse tensor cores, reducing metadata overhead, and minimizing on-chip decoding costs.

\item We design a SpMM GPU kernel for moderately unstructured sparse matrices that efficiently co-utilizes sparse tensor cores and CUDA cores. Its pipelined execution overlaps computation with memory access, thereby maximizing HBM bandwidth utilization.

\item We implement our approach for both the SpMM kernel and end-to-end LLM inference. On modern GPUs equipped with HBM, our method is the first to outperform dense matrix multiplication. It achieves up to 1.64× kernel-level speedup over SpInfer (EuroSys’25, Best Paper) and up to 1.41× end-to-end speedups over FlashLLM (VLDB’24).

\end{itemize}

%% file: text/background.tex
\subsection{Large Language Model and Model Pruning}

\begin{figure}[t]
\centering
\includegraphics[scale=0.65]{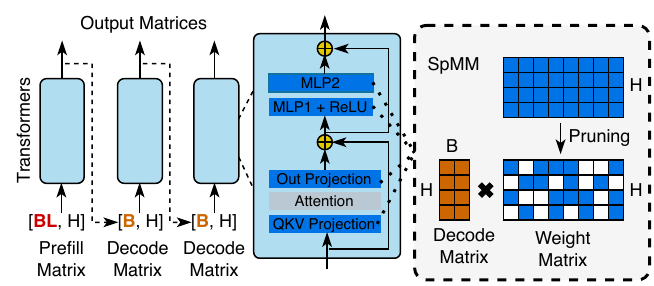}
\caption{The workflow of LLM inference with sparsity introduced by model pruning.}
\label{fig:llm}
\end{figure}

The inference of large language models (LLMs) consists of two phases: prefill and decode. As shown in Figure \ref{fig:llm}, the prefill phase processes the full input sequence to generate the first output token, with input shape $[\mathit{B}\mathit{L}, \mathit{H}]$. The decode phase then iteratively takes the previously generated token as input, handling a thinner matrix $[\mathit{B}, \mathit{H}]$ at each step. In both phases, matrix multiplication dominates computation, forming the core workload of LLM inference.

Model pruning, by systematically removing less important elements from weight matrices, directly reduces matrix size and computational cost. Since pruning targets the weight matrices directly, the sparsity pattern can be precomputed and stored in a specialized format ahead of time, eliminating the need for additional processing during model inference. This allows for faster inference execution.

In this work, we focus on LLMs using the sparsity patterns introduced by model pruning, which are orthogonal to quantization \cite{quant}, mixture-of-exports \cite{moe-1}, and sparse attention \cite{satten1} approaches.

\begin{figure}[t]
\centering
\includegraphics[scale=0.17]{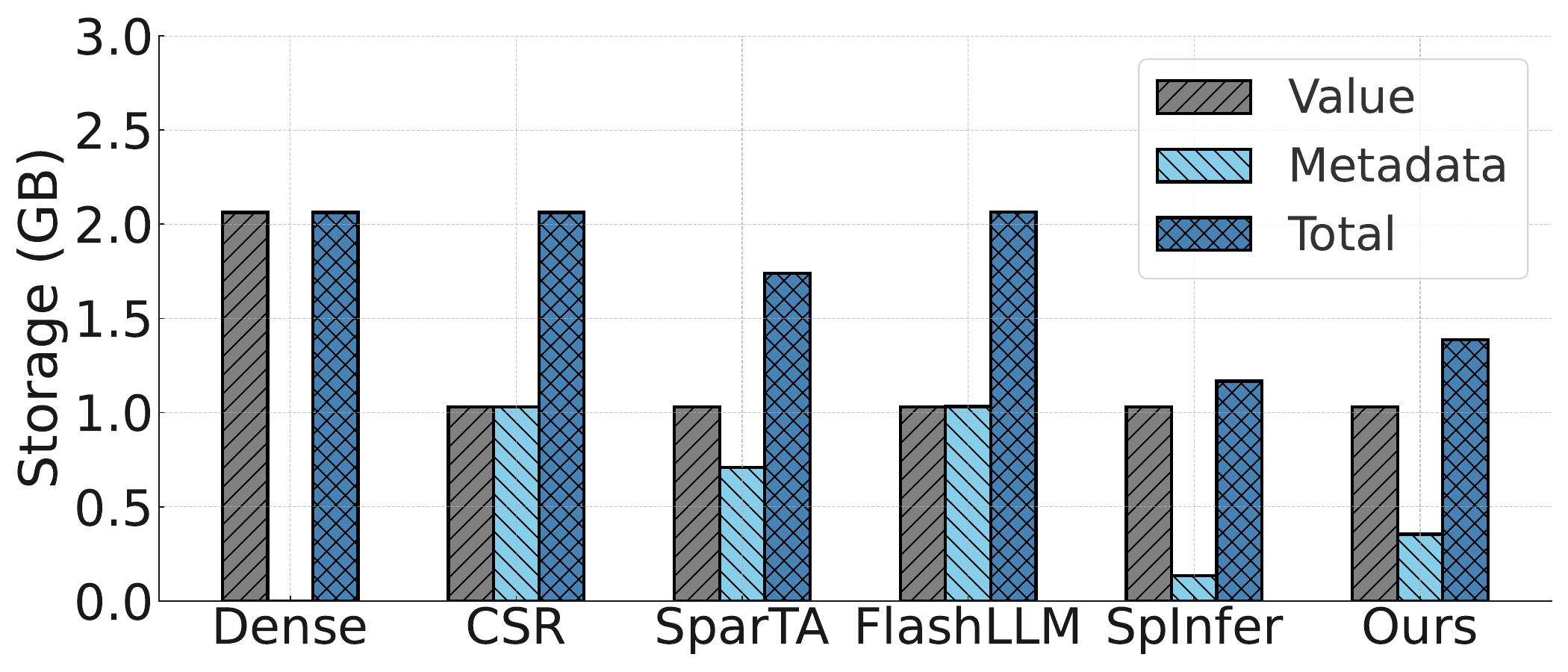}
\caption{Comparison of storage costs for different sparse matrix formats at 50\% sparsity level.}
\label{fig:storage}
\end{figure}

\begin{figure*}[t]
\centering
\includegraphics[scale=0.56]{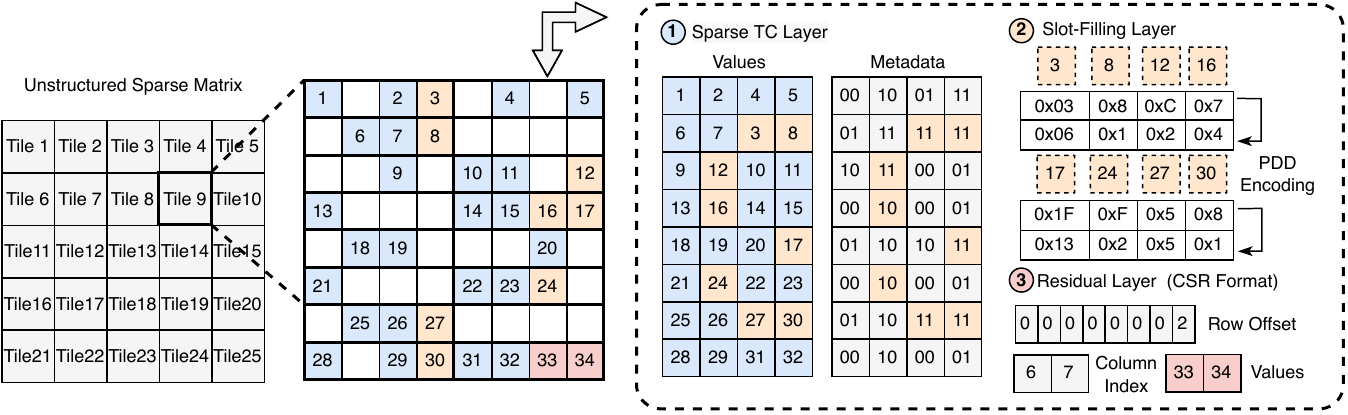}
\caption{Multi-layer storage format for sparse matrices with moderate unstructured sparsity.}
\label{fig:format}
\end{figure*}

\begin{figure}[t]
\centering
\includegraphics[scale=0.34]{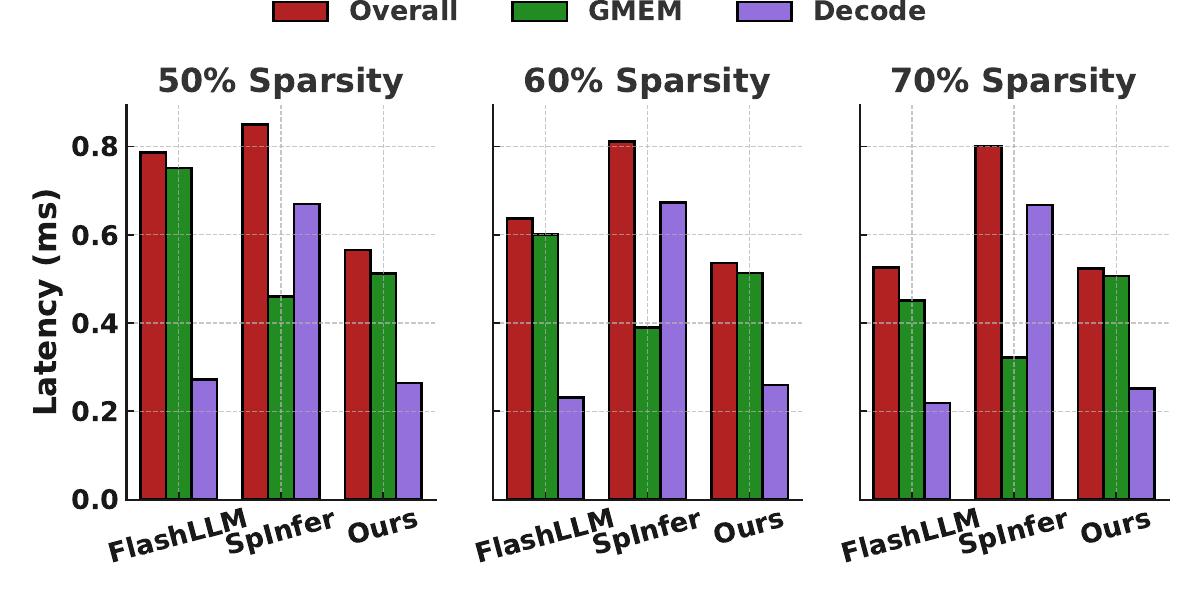}
\caption{Comparison of global memory access and on-chip decoding latency at different sparsity.}
\label{fig:gmem_rd}
\end{figure}

\vspace{-0.2cm}
\subsection{Modern GPU Architecture}
\label{sec:mgpu}

Modern GPUs are designed for massively parallel computation, making them highly effective for deep learning and large language model (LLM) inference. A GPU integrates thousands of CUDA cores capable of executing threads concurrently. To further accelerate AI workloads, modern GPUs incorporate tensor cores, specialized hardware units optimized for matrix operations. Dense tensor cores handle standard matrix multiplication, while sparse tensor cores exploit 2:4 structured sparsity to nearly double throughput.

Memory bandwidth is another key determinant. Modern GPUs adopt a hierarchical memory architecture consisting of global memory, an L2 cache, and fast on-chip shared memory. To further improve data transfer efficiency, recent architectures integrate high-bandwidth memory (HBM) as the global memory, offering higher throughput than traditional GDDR memory. This design enables rapid access to model weights required by modern AI workloads.

\vspace{-0.2cm}
\subsection{Shortcomings of Existing Work}

Extensive research efforts \cite{sparta, sputnik, highsc, tiled_csr, qgtc, sparRNN, sparsetir, tcgnn, spinfer} have focused on accelerating sparse matrix multiplication on GPUs. However, many GPU kernels are primarily designed for scientific applications \cite{heuristic, dtc, fast, adaptive, row}, where matrices are extremely sparse (exceeding 99\%). The state-of-the-art solutions designed for AI workloads include SparTA \cite{sparta}, FlashLLM \cite{flashllm}, and SpInfer \cite{spinfer}, each proposing a novel sparse matrix storage format with a GPU execution scheme.

SparTA \cite{sparta} partitions unstructured matrices into a 2:4 structured part and a residual unstructured part, accelerating them with cuSPARSELt \cite{cuSPARSELt} on sparse tensor cores and Sputnik \cite{sputnik} on CUDA cores, respectively. However, at moderate sparsity, many nonzeros failing to meet the 2:4 pattern move to the unstructured part, overloading CUDA core execution. FlashLLM \cite{flashllm} loads matrices sparsely but converts them to dense form for tensor-core execution. While this reduces global memory traffic, its CSL storage introduces heavy metadata overhead at moderate sparsity, where storage size approaches that of dense matrices (Figure \ref{fig:storage}).
SpInfer \cite{spinfer} employs a bitmap encoding that replaces explicit indices with single-bit nonzero indicators, improving memory access efficiency but introducing notable on-chip decoding overhead, particularly on GPUs with HBM, where the on-chip computation cannot be fully overlapped by global memory access (Figure \ref{fig:gmem_rd}).

%% file: text/storage.tex
\label{sec:storage}

\subsection{Design Goals}
\label{sec:storage_goal}

The design targets three requirements. First, it aims to reduce the overhead of positional metadata, which encodes the positions of non-zero elements in sparse matrices. Second, the format must be compatible with sparse tensor cores, which provide efficient hardware acceleration for structured sparse matrix operations on GPUs. Third, it is designed to reduce on-chip decoding overhead. Since metadata decoding is processed by the slower CUDA cores, excessive decoding complexity can become a major bottleneck.

\subsection{Multi-layer Storage Format for Matrices}

To achieve our design goals, we develop a multi-layer storage format for sparse matrices, as illustrated in Figure \ref{fig:format}. The format comprises three layers: the Sparse-TC layer, the Slot-Filling layer, and the Residual layer, each addressing a distinct aspect of sparsity.

\textbf{Sparse-TC Layer.} This layer extracts 2:4 structured sparsity patterns from the irregular distribution of non-zero elements, retaining two non-zero values from every four consecutive weights. Regions lacking sufficient non-zeros or containing surplus elements are deferred to the Slot-Filling layer. Because the extracted positions follow a fixed 4-element pattern, only 2 bits per value are needed to specify their position within the window, ensuring compact metadata storage and direct compatibility with modern sparse tensor cores without extra decoding.

\textbf{Slot-Filling Layer.} The Slot-Filling layer is designed to redistribute the surplus non-zero elements that exceed the capacity of a 2:4 window into the empty slots left unused in the Sparse-TC layer. Around 17.7\% of the total non-zero elements fall within this layer. Therefore, directly storing full row and column indices remains costly. To reduce this overhead, we adopt a parallel differential distance (PDD) encoding to represent the positional metadata of this layer, which captures the distances between successive redistributed elements. Based on the observed density of these elements in the matrix, most positions can be efficiently encoded using just 4 bits, which allows representing distances of up to 16. For example, in Figure \ref{fig:format}, the 3rd element originally located at position 3 is shifted to position 6, while another 4th element from position 11=3+8 is moved to position 7=6+1. For cases where the distance between two elements exceeds 16, we insert dummy points to break the span into smaller segments, ensuring that no individual distance exceeds 16. Dummy points account for 2.13\% of non-zero elements.

\textbf{Residual Layer.} The remaining non-zero elements that cannot be efficiently stored in either the Sparse-TC or Slot-Filling layers are handled by the Residual Layer, which uses a traditional sparse matrix format such as Compressed Sparse Row (CSR). Although traditional formats incur higher metadata overhead to store explicit row and column indices, the volume of data routed to this layer is minimal, typically less than 1\%. Thus, their metadata overhead has a negligible impact on overall efficiency.

%% file: text/kernel.tex

\subsection{Overall Workflow}

In general, our SpMM kernel leverages the cooperation between sparse tensor cores and CUDA cores to efficiently accelerate sparse matrix multiplication. The workflow includes: (1) Efficient Matrix Loading, where both the sparse matrix and the dense input matrix are loaded from global memory into on-chip memory; (2) CUDA Core Computation, where CUDA cores are used to decode parallel differential distance (PDD) encoded metadata from the Slot-Filling layer and process extreme sparse matrix in the Residual layer; and (3) Sparse Tensor Core Computation, where sparse tensor cores are used to accelerate structured matrix operations from the Sparse-TC layer. In addition, we implement a pipeline across sparse tensor cores, CUDA cores, and memory access, overlapping on-chip execution with global memory access to maximize resource utilization.

\begin{figure}[t]
\centering\includegraphics[scale=0.65]{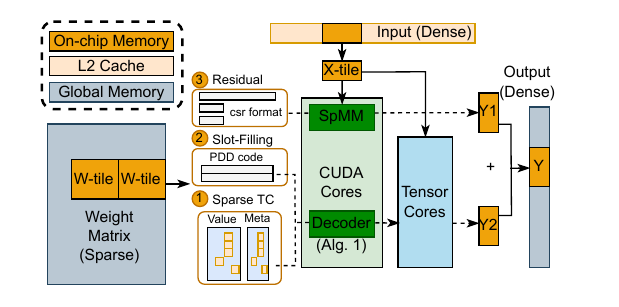}
\caption{The workflow of our SpMM kernel.}
\label{fig:spmm}
\end{figure}

\vspace{-0.2cm}
\subsection{Efficient Matrix Loading}
\label{sec:loading}

Efficiently loading both the sparse weight matrix and the dense input matrix from global memory into on-chip memory is key to high-performance sparse matrix multiplication. In LLM decoding, the dense input matrix is small,  typically of the size $B×H$. For example, $B = 32$, $H = 8\text{K}$ yields a matrix under 10\,MB, which fits in the L2 cache of most GPUs. We exploit this by caching the dense matrix in L2, so it is reused from cache instead of repeatedly fetched from global memory, allowing for faster access.

The sparse matrix represents the weight matrix in LLMs, and its size exceeds the capacity of the L2 cache. To address this, we adopt a tile-based loading strategy, where the matrix is partitioned into smaller tiles that are loaded iteratively. The tile size is based on the matrix shape supported by the tensor cores. Specifically, we use the \texttt{wgmma} instruction to perform matrix operations on the sparse tensor cores, which support computations in the format of $64 \times 32$. We select a tile size of $64 \times 128$ for four consecutive \texttt{wgmma} operations within the loop using the \texttt{unroll} pragma.

\begin{algorithm}[t]
\caption{PDD Decoding pseudo code}
\label{alg:pdd}
\begin{algorithmic}[1]
\Require The number of threads $T$. A sparse matrix in our storage format, including the structured sparse matrix $\mathbf{W_t}$ in the Sparse-TC layer, the PDD code $\mathbf{C} = [C_1, C_2, ..., C_T]$ with $T$-way parallelism. An empty structured sparse matrix $\mathbf{W_r}$.

\State $tid = \mathsf{GetThreadID()};\ \ \ \  code = \mathbf{C}[tid]$
\State {\color{blue}{// The first position is encoded using 16 bits.}}
\State $\mathsf{offset_t}$ = BitExtract($code, start=0, len=16$)
\State $\mathsf{offset_r}$ = BitExtract($code, start=16, len=16$)
\State $\mathbf{W_r}[\mathsf{offset_r}] = \mathbf{W_t}[\mathsf{offset_t}];$ $\ \ \ \ \mathbf{W_t}[\mathsf{offset_t}] = 0$ 

\For{$i = 0$ to (BitLen($code$) - $32$) / $8$}
    \State {\color{blue}{// Each subsequent position is obtained by adding the 4-bit differential distance to the former one.}}
    
    \State $\mathsf{offset_t}$ += BitExtract($code, start=32+8i+0, len=4$)
    \State $\mathsf{offset_r}$ += BitExtract($code, start=32+8i+4, len=4$)
    \State $\mathbf{W_r}[\mathsf{offset_r}] = \mathbf{W_t}[\mathsf{offset_t}];$ $\ \ \ \ \mathbf{W_t}[\mathsf{offset_t}] = 0$ 
\EndFor

\State \Return $\mathbf{W_t}$, $\mathbf{W_r}$
\end{algorithmic}
\end{algorithm}

\begin{figure}[t]
\centering\includegraphics[scale=0.5]{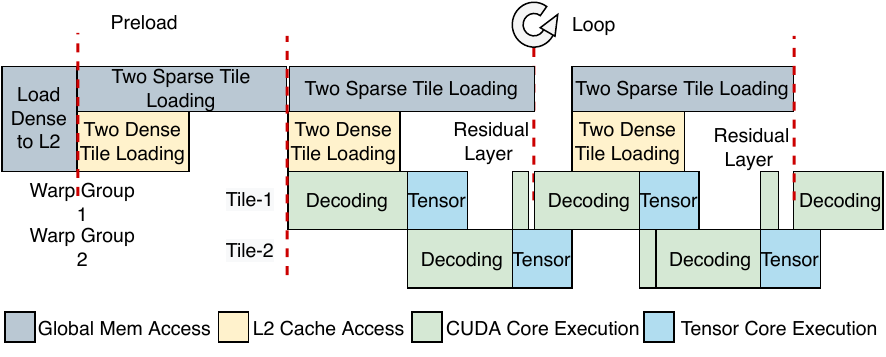}
\caption{A pipeline execution of our SpMM kernel.}
\label{fig:pipeline}
\end{figure}

\begin{figure*}[t]
\centering
\includegraphics[scale=0.49]{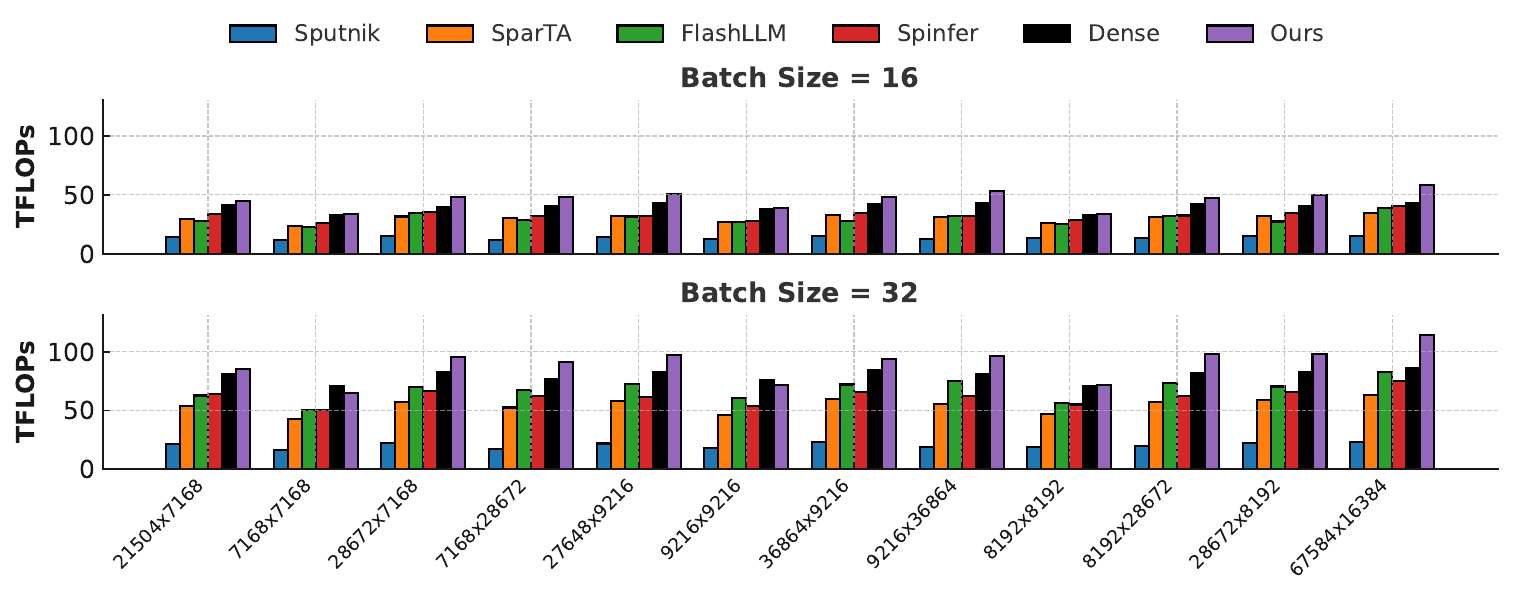}
\caption{SpMM kernel performance comparison across different matrix shapes at the 50\% sparsity level.}
\label{fig:tflops}
\end{figure*}

\vspace{-0.2cm}
\subsection{PDD decoding and Sparse TC Computation}
\label{sec:pdd}

The Slot-Filling layer redistributes surplus non-zero elements that exceed the capacity of a 2:4 structured sparsity window into empty slots. To minimize metadata overhead, the layer applies parallel differential distance (PDD) coding to compress the positional metadata for redistributed elements. During on-chip execution, CUDA cores are employed to decode the PDD metadata in parallel, detailed in Algorithm \ref{alg:pdd}. Each CUDA thread is assigned a portion of the PDD-encoded data, which contains relative offsets between element positions. The decoding process is executed as a parallel prefix sum over the delta values to recover the absolute positions. As the redistributed elements represent only about 17.7\% of the total, the decoding overhead does not pose a performance bottleneck.

Next, we employ the sparse tensor cores to process both the sparse matrix in the Sparse-TC layer. We leverage the \texttt{wgmma} instruction, introduced in NVIDIA's Hopper architecture, which extends the traditional \texttt{mma} instruction by enabling multiple warps to collaboratively perform larger and more efficient matrix operations. To maximize performance, we ensure that input operands are properly aligned and that the data follows the 2:4 structured sparsity format required by \texttt{wgmma}. Additionally, we utilize asynchronous execution between CUDA cores and sparse tensor cores.

\vspace{-0.2cm}
\subsection{Pipeline Design}
\label{sec:pipeline}

We develop a fine-grained pipeline to improve our GPU kernel for sparse matrix multiplication. The schematic representation of this pipeline is shown in Figure \ref{fig:pipeline}, which enables multi-level overlap across memory access, tensor core execution, and CUDA core computation, ensuring high resource utilization.

\textbf{Overlapping sparse and dense matrix loading.} As described in Section \ref{sec:loading}, we preload the entire dense input matrix into the L2 cache, leveraging its small size in LLMs. In contrast, the sparse weight matrix, which is much larger, remains in global memory and is fetched tile by tile. This hierarchical memory strategy allows concurrent access to the dense matrix in L2 and the sparse matrix from global memory, improving the throughput of matrix loading.

\textbf{Overlapping tensor and CUDA cores execution.} As described in Section \ref{sec:pdd}, our GPU kernel coordinates sparse tensor cores and CUDA cores: tensor cores accelerate structured matrix multiplications, while CUDA cores handle PDD decoding. Since these units run on separate hardware, their work can overlap. During matrix loading, we group two tiles and schedule two warp groups as in Figure \ref{fig:pipeline}, so that tensor cores compute the current tile while CUDA cores decode and redistribute the next tile, improving pipeline utilization. CUDA cores also process the extremely sparse Residual layer matrix, which contributes less than 1\% of the total workload. To avoid contention with tensor-core outputs, we use a separate on-chip buffer for the Residual results and perform a single reduction with the tensor-core results at the end of the pipeline.

\textbf{Overlapping memory access and on-chip execution} Our pipeline improves the overlap between memory access and on-chip execution by using a load-as-sparse, compute-as-sparse strategy. Sparsity is preserved end-to-end—from matrix loading and sparse metadata decoding to computation on sparse tensor cores and processing of the highly sparse residual matrix. As shown in Figure \ref{fig:gmem_rd}, this sparse-aware pipeline yields a more balanced workload between global memory access and on-chip execution than FlashLLM \cite{flashllm} and SpInfer \cite{spinfer}, enabling more effective overlap.

%% file: text/evaluation.tex

\subsection{Experimental Setup}
\label{sec:setup}

We evaluate GPU kernels for sparse matrix multiplication and end-to-end GPU inference of LLMs on NVIDIA H100 SXM5 GPUs, each equipped with 80 GB of HBM3 serving as global memory. All GPUs are interconnected via pairwise NVLink for high-bandwidth communication. Our code is compiled using GCC 12.3.0 and NVCC 12.6. We use the notation $M/K/N$ to denote matrix dimensions, where the sparse weight matrix has size $M \times K$, the dense input matrix has size $K \times N$, and the dense output matrix has size $M \times N$. Our work targets the decoding phase of LLM inference, where $N$ is the batch size, typically small, such as $N=16$ and $N = 32$.

\vspace{-0.2cm}

\subsection{Evaluating SpMM Kernel}
\label{sec:eval_spmm}

\begin{figure}[t]
\centering
\includegraphics[scale=0.115]{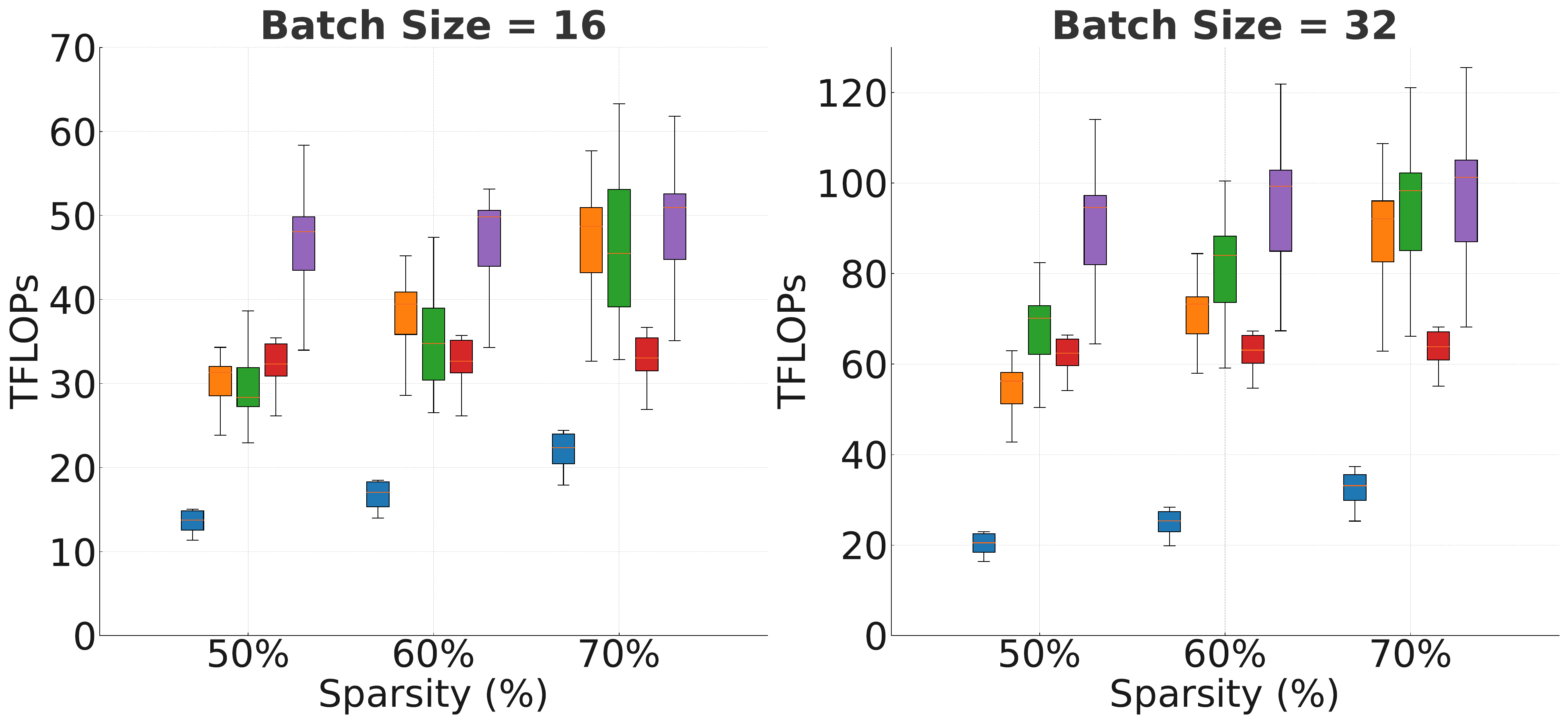}
\caption{SpMM kernel performance comparison across different sparsity levels. Blue: Sputnik \cite{sputnik}; Orange: SparTA \cite{sparta}; Green: FlashLLM \cite{flashllm}; Red: SpInfer \cite{spinfer}; Purple: Ours.}
\label{fig:boxplot}
\end{figure}

\textbf{Datasets and baselines}. We evaluate our SpMM kernel across a range of matrix shapes, primarily from LLMs such as OPT-30B and OPT-66B \cite{opt}, using batch sizes of 16 and 32. We measure the kernel throughput under sparsity levels of 50\%, 60\%, and 70\%. The baselines we compare against include Sputnik \cite{sputnik}, SparTA \cite{sparta}, FlashLLM \cite{flashllm}, SpInfer \cite{spinfer}. Sputnik is a state-of-the-art using CUDA cores. All other baselines are accelerated by tensor cores.

\textbf{Results}. As shown in Figure \ref{fig:tflops}, at the 50\% sparsity level, our SpMM kernel significantly outperforms all baselines. Compared to Sputnik, which is based on CUDA cores, our kernel achieves up to a 5.29× speedup and an average 3.92× improvement in throughput. For other baselines, including SparTA, FlashLLM, and SpInfer, which all leverage tensor core acceleration, our method still delivers substantial gains, achieving speedups of up to 1.82×, 1.81×, and 1.64×, respectively. Our work is the first to outperform dense matrix multiplication on modern GPUs equipped with HBM. 

As illustrated in Figure \ref{fig:boxplot}, we evaluate performance across sparsity levels ranging from 50\% to 70\%. At 60\% sparsity, our method achieves average speedups of 1.29×, 1.26×, and 1.47× over SparTA, FlashLLM, and SpInfer. Even at 70\% sparsity, beyond the primary optimization target of our kernel, our approach consistently maintains its advantage and outperforms all baseline methods.

\subsection{Evaluating End-to-end LLM Inference}
\label{sec:eval_llm}

\begin{figure}[t]
\centering
\includegraphics[scale=0.15]{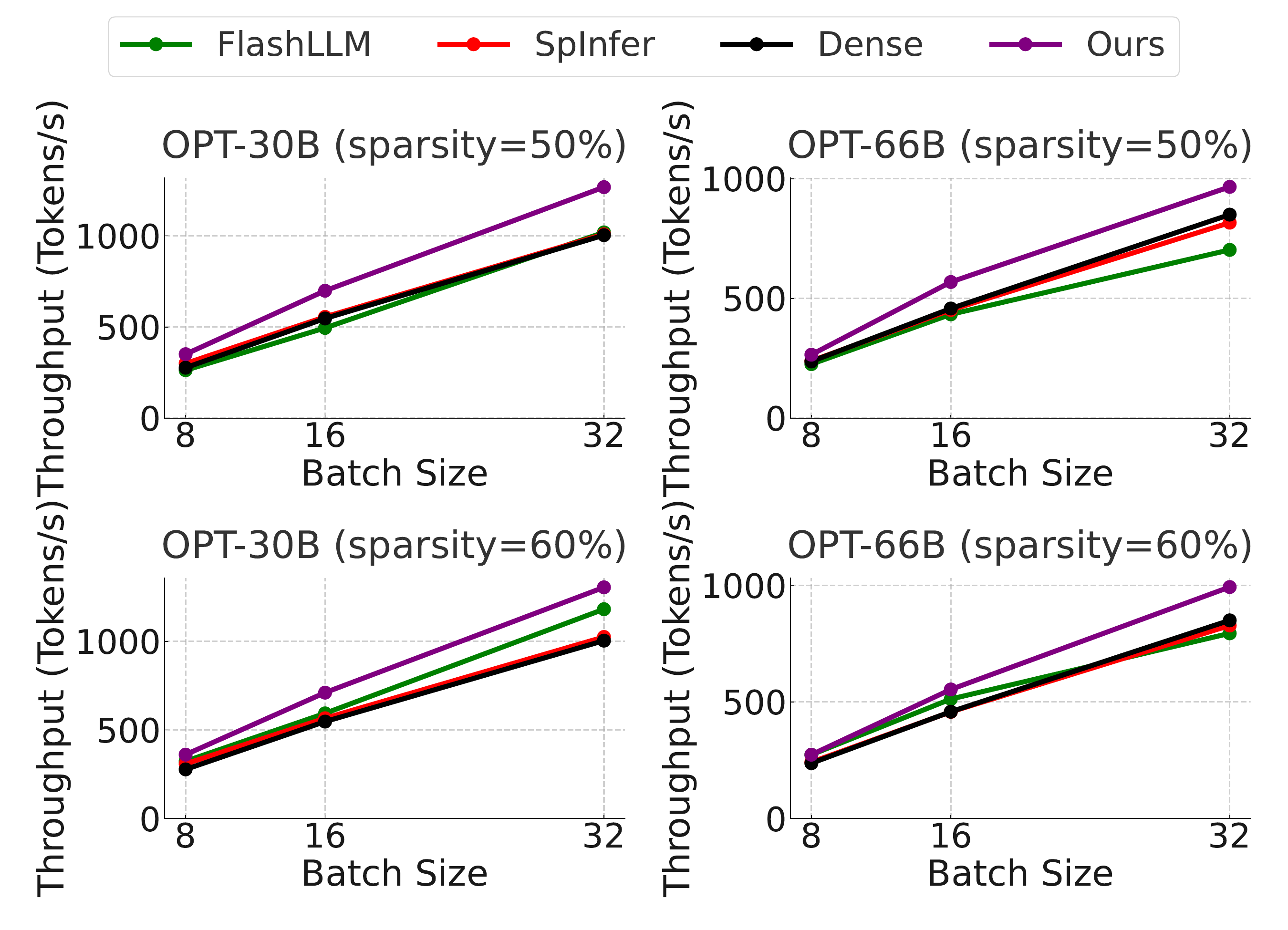}
\caption{End-to-end inference throughput per GPU.}
\label{fig:token}
\end{figure}

\textbf{Settings}. We perform evaluation using the OPT-30B and OPT-66B models. Model pruning is performed using the advanced Wanda algorithm~\cite{wanda}. Experiments are conducted with batch sizes of 8, 16, and 32. For OPT-66B inference, we use a 2-GPU configuration, except for FlashLLM~\cite{flashllm}, which requires 4 GPUs due to its large memory footprint. In all experiments, the input sequence length is set to 512, and the output sequence length is fixed at 512.

\textbf{Performance}. As shown in Figure~\ref{fig:token}, at the 50\% sparsity level, our method outperforms all baselines across both the OPT-30B and OPT-66B models. For example, on OPT-30B with a batch size of 16, our approach achieves a 41.4\% higher throughput than FlashLLM and 26.1\% higher than SpInfer. For OPT-66B, our method improves throughput by up to 37.6\% over FlashLLM and 20.2\% over SpInfer. At 60\% sparsity level, our method achieves throughput improvements of up to 19.7\% over FlashLLM and 27.4\% over SpInfer. The speedup of end-to-end LLM inference cannot be same as the kernel-level improvements, because it involves not only matrix multiplications but also other components such as activation functions.

\begin{figure}[t]
\centering
\includegraphics[scale=0.17]{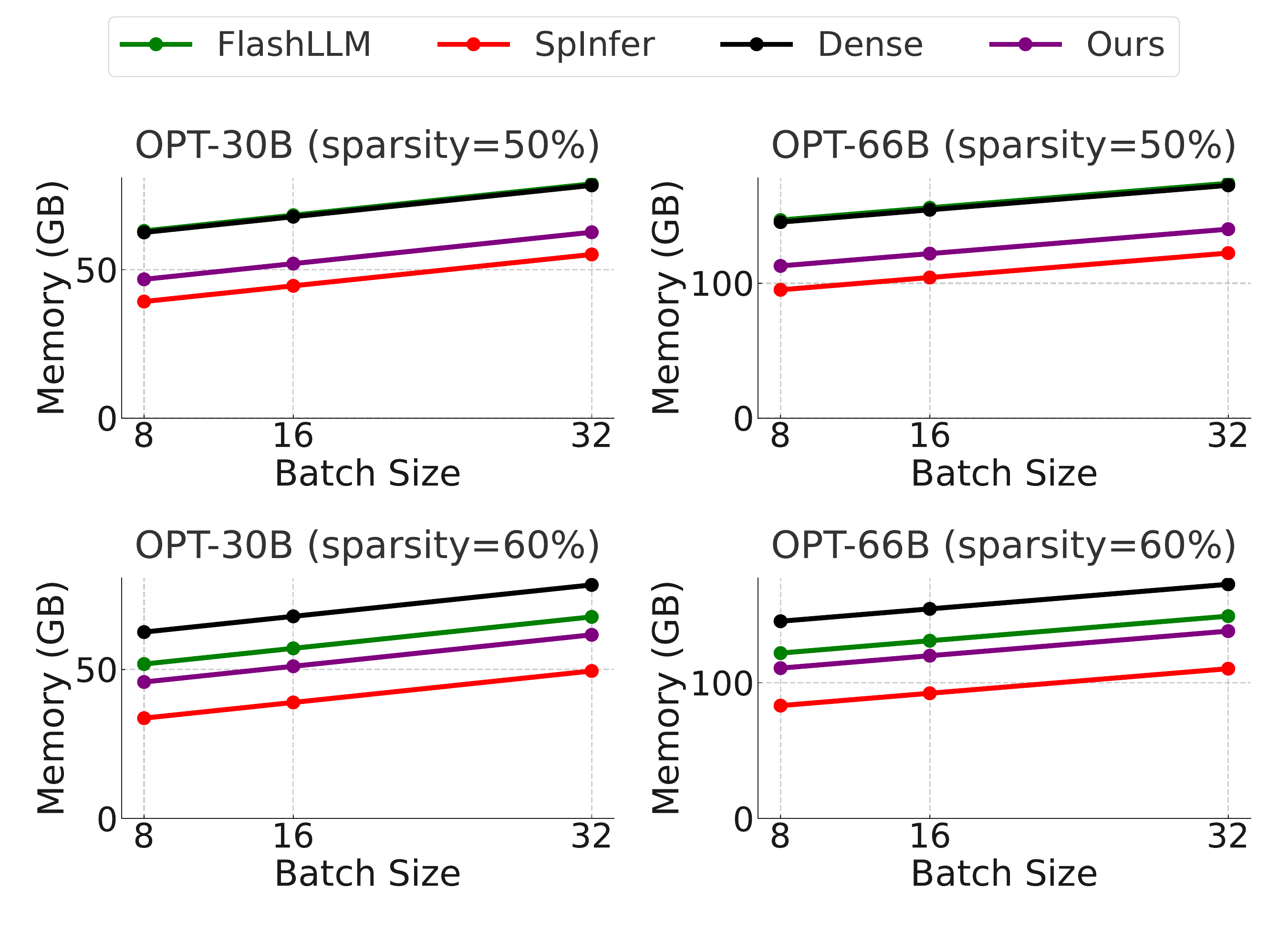}
\caption{Peak GPU memory usage in LLM inference.}
\label{fig:memory}
\end{figure}

\textbf{Peak Memory}. As shown in Figure \ref{fig:memory}, our method consistently reduces memory usage compared to FlashLLM and the dense format. For example, at 50\% sparsity on the OPT-30B model with batch size 16, our approach reduces global memory usage by approximately 21.4\% compared to the dense baseline. While our method does incur slightly higher memory usage than SpInfer, this is expected due to the latter's use of bitmap encoding for metadata.

\textbf{Storage Format Conversion}. Conversion from a dense format to our sparse format requires only a one-time cost. This eliminates the need for repeated processing during inference. On an Intel Xeon Platinum 8460Y+ CPU, converting all matrices takes 41.1 seconds for the OPT-30B model and 92.6 seconds for the OPT-66B model.

%% file: text/limitation.tex
\begin{figure}[t]
\centering
\includegraphics[scale=0.12]{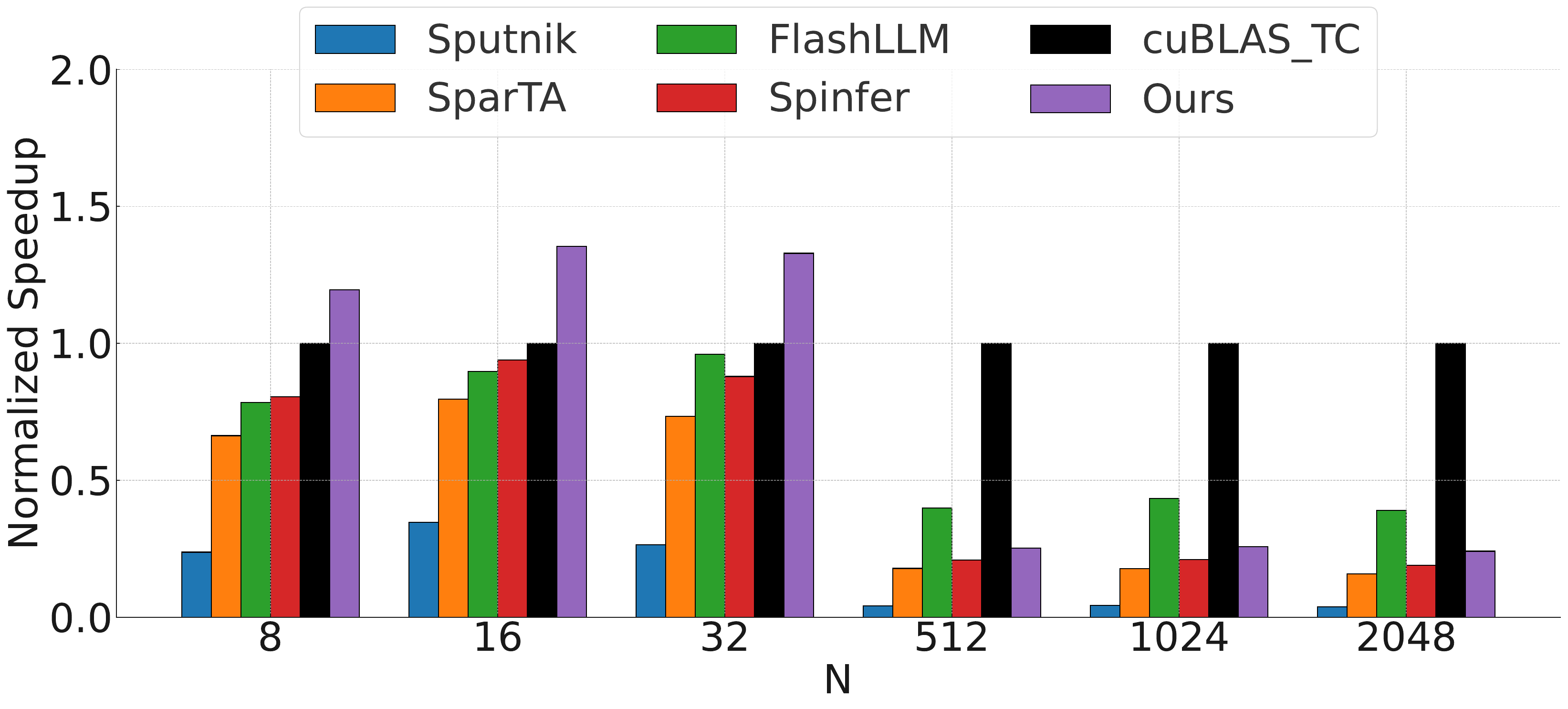}
\caption{Performance comparison of SpMM kernels under small and large N settings at 50\% sparsity.}
\label{fig:var_n}
\end{figure}

Our scheme does not outperform FlashLLM \cite{flashllm} under high sparsity. The limitation arises from the storage format, as most regions require zero padding while few contain surplus non-zeros, rendering our redistribution strategy ineffective. As shown in Figure \ref{fig:var_n}, our method, similar to SpInfer \cite{spinfer} and FlashLLM \cite{flashllm}, is constrained when $N$ becomes large during the prefill phase. In this compute-bound regime, our design, optimized for memory-bound decoding with small $N$, runs slower than cuBLAS \cite{cublas}. Nevertheless, the increasing adoption of architectures that decouple the prefill and decode phases \cite{exegpt, splitwise, mooncake} makes our decode-phase optimization suitable for scalable deployment. Our implementation targets GPUs equipped with sparse tensor cores and high-bandwidth memory (HBM). These features are now widely available in modern GPUs, such as NVIDIA Ampere \cite{Ampere} and Hopper \cite{Hopper} (A100, H100), as well as AMD’s MI300 series based on the CDNA3 architecture \cite{CDNA}. The widespread availability of such hardware enables practical deployment across diverse platforms.

%% file: text/conclusion.tex
\label{sec:conclusion}

In this paper, we propose a method to accelerate GPU inference for LLMs with sparse weight matrices. We begin by introducing a three-layer storage format designed to reduce metadata overhead, ensure compatibility with sparse tensor cores, and minimize on-chip decoding costs. Additionally, we design a GPU kernel that accelerates SpMM using an execution pipeline that maximizes memory bandwidth utilization through effective overlap of on-chip execution and global memory access. The evaluation results show that our scheme has a considerable speedup over other implementations.


\section*{Acknowledgments}

This work was supported by the CRPO under WBS A-8004052-01-00, the Tier 2 grant MOE-T2EP20125-0015, in part by the National Key Research and Development Program of China No. 2023YFB4404401, in part by the National Natural Science Foundation of China under Grant No. 92373205, in part by the Key Researchand Development Program of Zhejiang Province under Grant No. 2025C02103.